\documentclass[NumberedRefs,turnofflinenumbers]{JASA-EL}
\usepackage{dsfont}
\begin{document}

\title[Source ranging using a neural network]{Sound source ranging using a feed-forward neural network with fitting-based early stopping}
\author{Jing Chi}
\email{ytytcj110@163.com}
\author{Xiaolei Li}
\email{lxl\_ouc@outlook.com}
\correspondingauthor
\author{Haozhong Wang}
\email{coolice@ouc.edu.cn}
\author{Dazhi Gao}
\email{dzgao@ouc.edu.cn}
\correspondingauthor
\affiliation{Department of Marine Technology, Ocean University of China, Qingdao 266100, China}
\author{Peter Gerstoft}
\email{pgerstoft@ucsd.edu}
\affiliation{Scripps Institution of Oceanography, University of California San Diego, La Jolla, California 92093-0238, USA}

\date{\today}
\preprint{Li et al., JASA-EL}  

\begin{abstract}
 When a feed-forward neural network (FNN) is trained for source ranging in an ocean waveguide, it is difficult evaluating the range accuracy of the FNN on unlabeled test data. A fitting-based early stopping (FEAST) method is introduced to evaluate the range error of the FNN on test data where the distance of source is unknown. Based on FEAST, when the evaluated range error of the FNN reaches the minimum on test data, stopping training, which will help to improve the ranging accuracy of the FNN on the test data. The FEAST is demonstrated on simulated and experimental data.
\end{abstract}


\maketitle


\section{\label{sec:1} Introduction section}

Matched field processing (MFP) \cite{Bucker_1, Tolstoy_2, Baggeroer_3, Gingras_a1, Mecklenbrauker_a2, Debever_a3} for source localization has been developed for many years. It can have limited performance due to its sensitivity to the mismatch between model-generated replica fields and measurements. With the development of machine learning, source localization methods based on machine learning have been revived \cite{Niu_8, Niu_23, Wang_24, Ferguson_25, Huang_9}. As early as 1991, Steinberg et al.\cite{Steinberg_7} applied perceptrons for source localization in a homogeneous medium. Recently, Niu et al. \cite{Niu_8, Niu_23} performed ship ranging using a feed-forward neural network (FNN) trained on experimental data. Besides, a regression neural network (NN) \cite{Wang_24} and a convolutional NN \cite{Ferguson_25} are also trained on experimental data for underwater source ranging. Although a NN can be trained on experimental data, because of the difficulty to obtain amounts of ocean acoustic experimental data containing distance labels, it is cumbersome to train a NN on experimental data to realize sound source ranging in an ocean waveguide. Considering the rarity of experimental data, Huang et al. \cite{Huang_9} combined simulation data in close environments to train a deep NN for source localization. However, because of the space-time variation of the ocean waveguide environment, even if the training data includes both simulation and experimental data,  the test data is often different from the training data due to the difference of the environment. Therefore, the NN with the minimum ranging error on training data may not reach the minimum ranging error on testing data. If the distance of sound source of partial test data is known, then this part of the data can be used as validation data with the source distance as labels, and early stopping  \cite{Garvesh_20, Prechelt_21} can be used to improve the ranging accuracy of the NN in the test data.

Early stopping is a form of regularization based on choosing when to stop running an iterative algorithm and is usually used to enhance generalization performance of NN and to fight overfitting \cite{Garvesh_20, Prechelt_21}. Generalization performance means small error on examples not seen during training. Validation error, which is the average error of NN on validation data and is computed by labeled validation data, is chosen as the criterion of whether the NN stops training in early stopping method \cite{Garvesh_20, Prechelt_21}. During the training process, when validation error reaches the minimum, stop training. Thus the validation error on validation data is reduced by early stopping.
Generally, however, test data do not contain labels and cannot be used as validation data. In this case, people cannot use early stopping to improve the ranging accuracy of the NN in the test data. If the ranging error of the NN in the test data can be evaluated, it can be used as the criterion of whether the NN stops training, so as to optimize the ranging accuracy of the NN in the test data.

In this paper, a FNN is trained on simulation data to realize source ranging in an ocean waveguide. Different from \cite{Niu_8, Niu_23, Ferguson_25, Huang_9} and early stopping method, the evaluated ranging error of the FNN on test data where the distance of source is unknown is used as the criterion of whether the FNN stops training. To evaluate the ranging error of the FNN on test data, a method called fitting-based early stopping (FEAST) is introduced. Assuming that the track of an underwater source satisfies a known parameterized function, the FEAST evaluates the ranging error of the FNN by parameter fitting. The FEAST is demonstrated on simulated and experimental data.


\section{\label{sec:2} Simulation data preparation, FNN architecture and learning parameters}
\begin{figure}[ht]
\includegraphics[height=0.4\textheight]{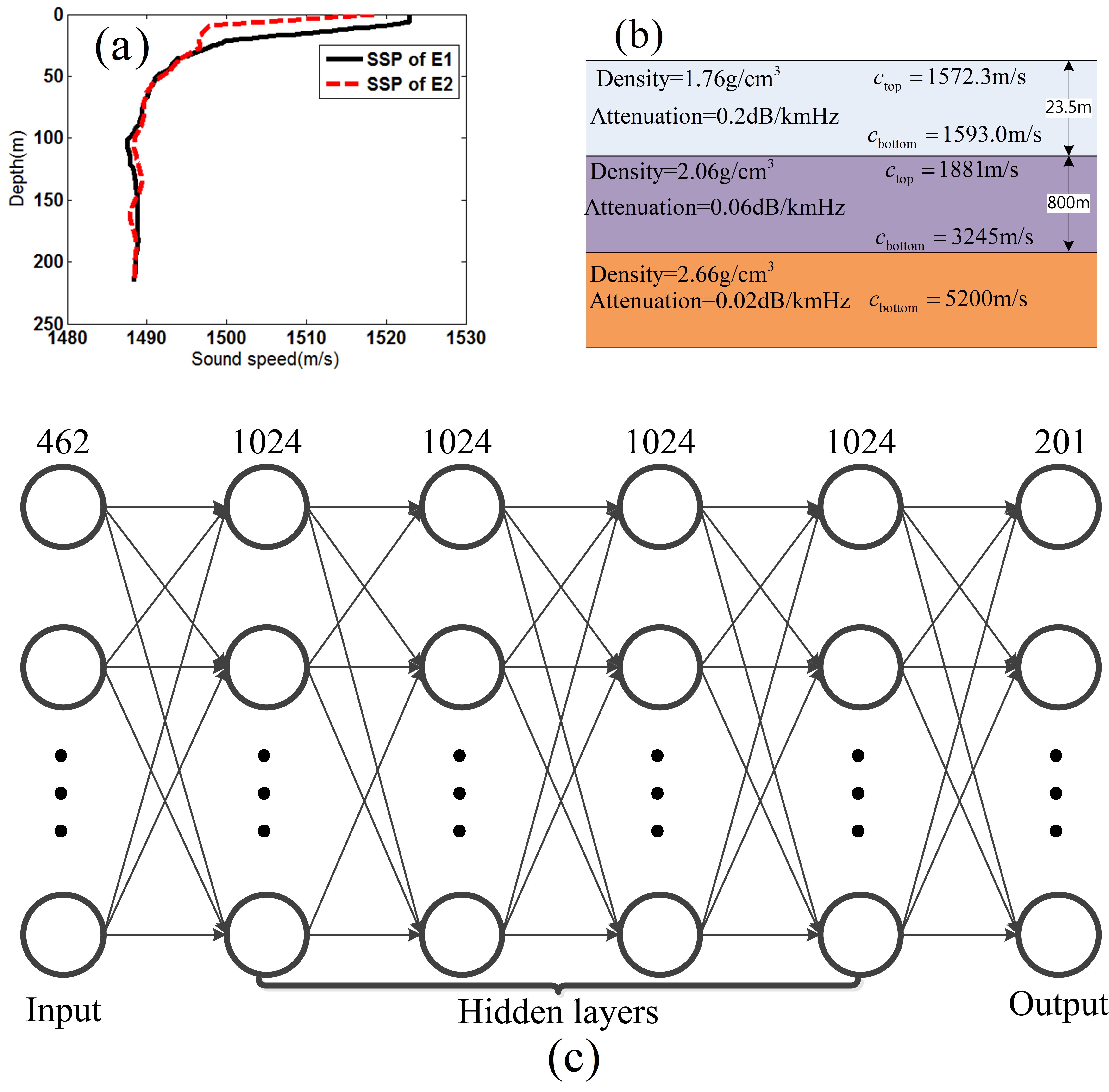}
\caption{(Color online) \label{fig1}{(a) Sound speed profile (SSP). (b) Seabed parameters. (c) Architecture of the FNN with 1024 neurons in each hidden layer, 462 in the input layer and 201 in the output layer.}}
\end{figure}

It will be useful to introduce the parameters used for simulation. Let E1 represent an range independent ocean waveguide which will be used for modeling the training data. The parameters of E1 are given by the S5 event in the SWellEx-96 experiment \cite{website_12}. The sound speed profile (SSP) and the seabed parameters of E1 are shown in Fig. \ref{fig1} (a) and (b).  The vertical line array (VLA) had 21 elements from 94.125--212.25 m in depth. Let E2 represent a range independent ocean waveguide which is used for modeling the test data. Except for the SSP, see Fig. \ref{fig1} (a), the parameters of E2 are the same as E1.

\begin{figure}[ht]
\includegraphics[height=0.35\textheight]{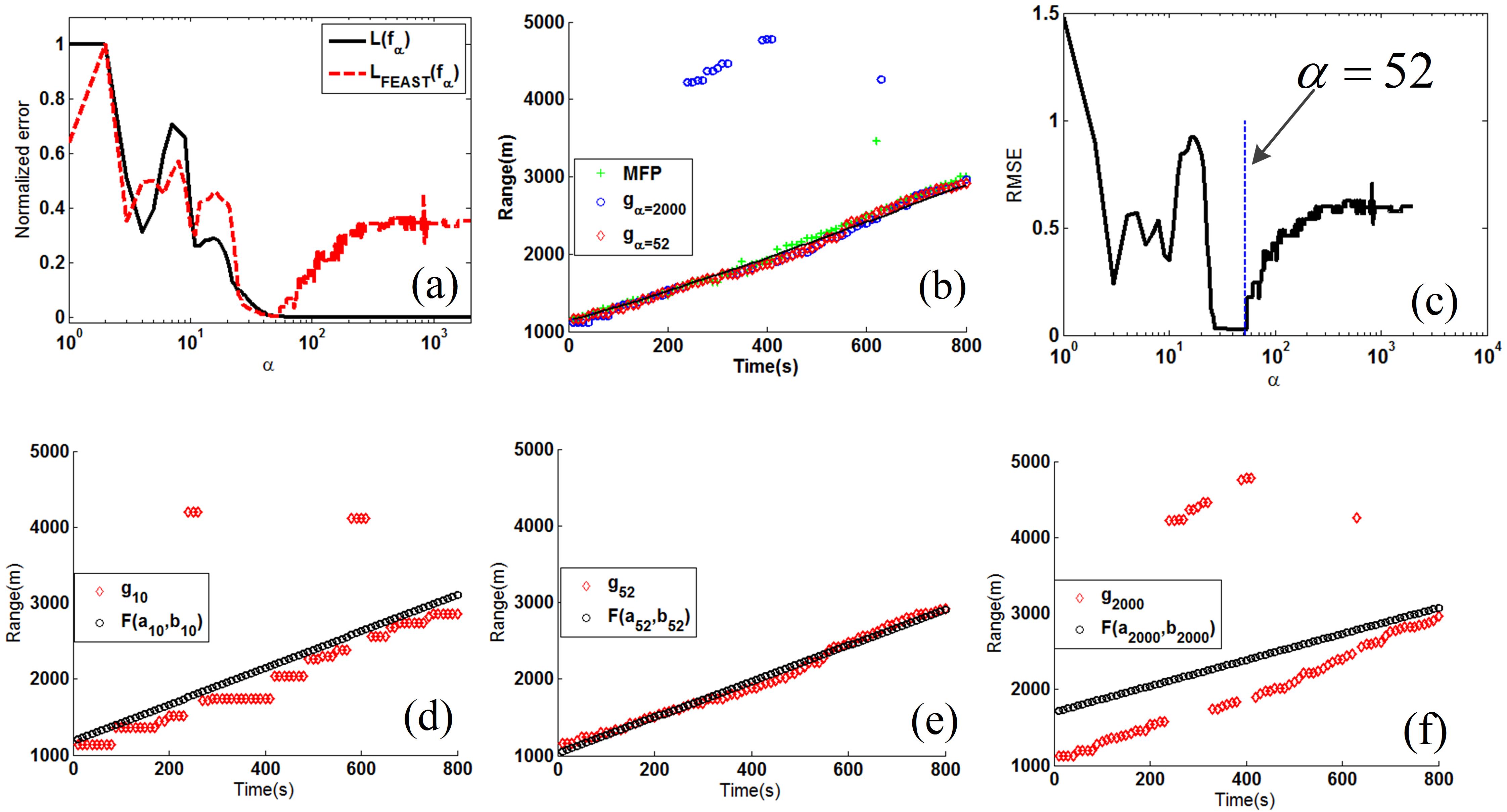}
\caption{(Color online)\label{fig2} Simulated data. {(a) The loss functions $L(\alpha)$ (solid) and $L_{\mathrm{FEAST}}(\alpha)$ (dashed). $L_{\mathrm{FEAST}}(\alpha)$ reaches the minimum when $\alpha=52$. (b) $g_{52}(\hat{\mathbf{x}}_i)$ (diamonds), $g_{2000}(\hat{\mathbf{x}}_i)$ (circles) and the range from MFP (crosses), and the real range of the source (solid). (c) Relative mean square error (RMSE) for $g_{\alpha}(\hat{\mathbf{x}}_i)$. $g_{\alpha}(\hat{\mathbf{x}}_i)$ (diamonds) and $F(\hat{\mathbf{x}}_i,a_\alpha,b_\alpha)$ (circles) for (d) $\alpha=10$, (e) $\alpha=52$ and (f) $\alpha=2000$.}}
\end{figure}

The simulated training and test data set are prepared as follows. Selecting a domain $D$ in E1, which is 1100--5000 m in range from the VLA and 1--30 m below the sea surface, a training set containing $12,000$ samples is constructed by choosing $12,000$ source locations in $D$ uniformly. Let $\mathbf{p}=[p_1,\cdots,p_{21}]^{\mathrm{T}}$ represent the acoustic signal received by the VLA when a 232 Hz point source is in the $D$, computed by Kraken \cite{kraken}. Then the input of FNN can be constructed by vectorizing the normalized sample covariance matrix $\mathbf{C}=\mathbf{p}\mathbf{p}^{\mathrm{H}}/\|\mathbf{p}\|^2_2$, see \cite{Niu_8}. Considering the Hermite symmetry of $\mathbf{C}$ and the fact that $\mathbf{C}$ is a complex matrix, $\mathbf{C}$ contains $11\times21\times2$ independent real numbers, which make up $\mathbf{x}\in\mathds{R}^{462\times1}$, the input of the FNN.
The label in the training set is obtained by dividing $D$ into $201$ parts uniformly in range direction and encoding the distance information in a 201-dimensional vector $\mathbf{y}\in \mathds{R}^{201\times1}$. If a source is in the $m\mathrm{th}$ part of $D$, the $m\mathrm{th}$ element of $\mathbf{y}$ is 1, all others are 0. The test set is generated by a moving $232$ Hz point source positioned 9 m below the sea surface and leaves the VLA in E2 at uniform velocity; see the black solid line in Fig. \ref{fig2} (b). The VLA records data at every 10 s and records 80 sets of data. When recording data, the moving point source is considered static. Then the test data which contain 80 samples are constructed in the same way as the training data except that the test data contains no labels. The differences between training and test data are mainly caused by environmental differences.

A four-hidden-layered FNN with 1024 neurons in each hidden layer is used, see Fig. \ref{fig1} (c). The input layer has 462 neurons, and the output layer has 201 neurons. Sigmoid function is selected as the activation function of the neurons in the hidden layers and softmax function in the output layer. The FNN is trained on TensorFlow with a learning rate of 0.0005 and the cross-entropy loss function is chosen to optimize the FNN. The cross-entropy loss function $L(\alpha)$ is:
\begin{equation}
L(\alpha)=-\frac{1}{N}\sum_{i=1}^N[\mathbf{y}_i^\mathrm{T}\ln\mathbf{f}_\alpha(\mathbf{x}_i)+(\mathbf{1}-\mathbf{y}_i)^\mathrm{T}\ln(\mathbf{1}-\mathbf{f}_\alpha(\mathbf{x}_i)],
\label{eq1}
\end{equation}
where $\alpha\in \mathds{N}$ represents epoch which is a measure of number of iterations in training, $N$ is the number of training samples, $\{\mathbf{x}_i,\mathbf{y}_i\}$ represents the $i\mathrm{th}$ training samples, $\mathbf{x}_i\in\mathds{R}^{462\times1}$ is the input to FNN, $\mathbf{y}_i\in\mathds{R}^{201\times1}$ is the label of $\mathbf{x}_i$,  $\mathbf{f}_{\alpha}:\mathds{R}^{462\times1}\rightarrow\mathds{R}^{201\times1}$ represents the trained FNN when epoch is $\alpha$,
``T'' is to transpose, and $\mathbf{1}\in\mathds{R}^{201\times1}$ is a vector with all elements 1. The $m\mathrm{th}$ element of $\mathbf{f}_\alpha (\mathbf{x}_i)$ represents the probability of a source in the $m\mathrm{th}$ part of $D$. The maximum of $\mathbf{f}_\alpha(\mathbf{x}_i)$ represent the likely source position and the source--VLA range is expressed by $g_\alpha(\mathbf{x}_i)$.
When $\mathbf{f}_\alpha(\mathbf{x}_i)=\mathbf{y}_i$, $l(\mathbf{f}_\alpha(\mathbf{x}_i),\mathbf{y}_i)$ has a minimum of 0. Fig. \ref{fig2} (a) shows $L(\alpha)$.

\section{\label{sec:3} Basic idea of FEAST}

Although the test data do not contain labels, the performance of a FNN on the test data can be evaluated in some situations. If the expected output of a FNN on the test data satisfies a known parameterized function $F(\hat{\mathbf{x}}({t}_i),\mathbf{\Theta})$,
where $\mathbf{\Theta}$ represents unknown fitting parameters, $\hat{\mathbf{x}}_i=\hat{\mathbf{x}}({t}_i)$ is the $i$th test data and ${t}_i$ is a known parameter,
\[
\min_{\mathbf{\Theta}}{\sqrt{\frac{1}{M}\sum_{i=1}^M[g_\alpha(\hat{\mathbf{x}}_i)-F(\hat{\mathbf{x}}_i, \mathbf{\Theta})]^2}}
\]
is used to evaluate the performance on the test data, where $M$ is number of test data.
Take a moving source from a a VLA as the example in this paper. Generally, the distance between a moving source and the VLA is a simple curve in the time-distance plane, which can be fitted by polynomials of finite order. For example, if the source moves from the VLA at constant speed, the function is the first-order polynomial on time, and if the source moves from VLA at constant acceleration, the function is a second-order polynomial.
For simplicity, only a source moving at constant speed is considered, thus $F(\hat{\mathbf{x}}_i,\mathbf{\Theta})=F(\hat{\mathbf{x}}(t_i),a,b)=a t_i+b$, where $t_i$ is the $i\mathrm{th}$ time instance and $\mathbf{\Theta}=[a,b]$. Define one loss function as
\begin{equation}
L_{\mathrm{FEAST}}(\alpha)=L(\alpha)+\lambda\sqrt{\frac{1}{M}\sum_{i=1}^M[g_\alpha(\hat{\mathbf{x}}_i)-F(\hat{\mathbf{x}}_i,a_\alpha,b_\alpha)]^2},
\label{eq6}
\end{equation}
where $L(\alpha)$ is defined in Eq. (\ref{eq1}),
\begin{equation}
\{a_\alpha,b_\alpha\}=\arg\min_{a,b}\sum_{i=1}^M[g_\alpha(\hat{\mathbf{x}}_i)-F(\hat{\mathbf{x}}_i,a_\alpha,b_\alpha)]^2,
\label{eqa1}
\end{equation}
and $\lambda\in\mathds{R}$ is a regularization parameter. Here
\begin{equation}
\lambda=\frac{\sqrt{M}\max{[L(\alpha)]}}{\max{\big[\sqrt{\sum_{i=1}^M[g_\alpha(\hat{\mathbf{x}}_i)-F(\hat{\mathbf{x}}_i,a_\alpha,b_\alpha)]^2}\big]}}
\label{eqa2}
\end{equation}
to make the maximum value of the two terms on the right side of Eq. (\ref{eq6}) equal to each other, and generally, the two terms reach their maximum values at small $\alpha$ (the two terms reach their maximum values in $\alpha<10$ in this paper).
The first term on the right side of Eq. (\ref{eq6}) is the loss function defined on the training data, which aims to avoid that the initialization result of the FNN satisfies the parameterized function; the second term computes the difference between $g_\alpha(\hat{\mathbf{x}}_i)$ and the parametric model of known form $F(\hat{\mathbf{x}}_i,a_\alpha,b_\alpha)$ and evaluates the ranging error of the FNN on the test data. When $L_{\mathrm{FEAST}}(\alpha)$ reaches the minimum or converges, stop training the FNN. Note that the training process of the FNN is completed by optimizing $L(\alpha)$, and $L_{\mathrm{FEAST}}(\alpha)$ just indicates when to stop. Because it is necessary to calculate $L_{\mathrm{FEAST}}(\alpha)$ by fitting parameters $\{a_\alpha,b_\alpha\}$ and $L_{\mathrm{FEAST}}(\alpha)$ reaches its minimum before $L(\alpha)$, this method is called fitting-based early stopping (FEAST). Not only in FNN, the FEAST is used in other types of neural network to improve ranging accuracy on test data.

To demonstrate FEAST, the test data prepared in Sec. \ref{sec:2} is used to calculate $L_{\mathrm{FEAST}}(\alpha)$. Fig. \ref{fig2} (a) shows $L_{\mathrm{FEAST}}(\alpha)$ which has minimum at $\alpha=52$. In order to facilitate the understanding of FEAST, Fig. \ref{fig2} (d)-(f) show $g_\alpha(\hat{\mathbf{x}}_i)$ and $F(\hat{\mathbf{x}}_i,a_\alpha,b_\alpha)$ at different $\alpha$, and one find that $g_\alpha(\hat{\mathbf{x}}_i)$ and $F(\hat{\mathbf{x}}_i,a_\alpha,b_\alpha)$ are similar when $L_\mathrm{FEAST}(\alpha)$ reaches its minimum.  Fig. \ref{fig2} (b) indicates that $g_{52}(\hat{\mathbf{x}}_i)$ is close to the true range in the test data. For comparison, Fig. \ref{fig2} (b) also shows $g_{2000}(\hat{\mathbf{x}}_i)$ and the range from MFP where the ocean waveguide environment used in MFP is E1. Except for the points near 620 s, the range from MFP and $g_{52}(\hat{\mathbf{x}}_i)$ are almost the same and slightly deviate from the true source distance which is caused by the difference between E1 and E2. However, the ranging results of $g_{2000}(\hat{\mathbf{x}}_i)$ have larger derivations. Define the relative mean square error (RMSE) for ranging:
\begin{equation}
\mathrm{RMSE}=\sqrt{\frac{1}{M}\sum_{m=1}^M\frac{[Rp(\hat{\mathbf{x}}_i)-Rt(\hat{\mathbf{x}}_i)]^2}{Rt(\hat{\mathbf{x}}_i)^2}},
\end{equation}
where $Rp(\hat{\mathbf{x}}_i)$ and $Rt(\hat{\mathbf{x}}_i)$ are the predicted range and the ground truth range corresponding to $\hat{\mathbf{x}}_i$. Fig. \ref{fig2} (c) gives the RMSE of $g_{\alpha}(\hat{\mathbf{x}}_i)$. One can find that when $\alpha=52$, the value of RMSE (0.0252) is closed to the minimum (0.0251), which verifies the FEAST.

\section{\label{sec:4} Experimental results}

\begin{figure}[ht]
\includegraphics[height=0.35\textheight]{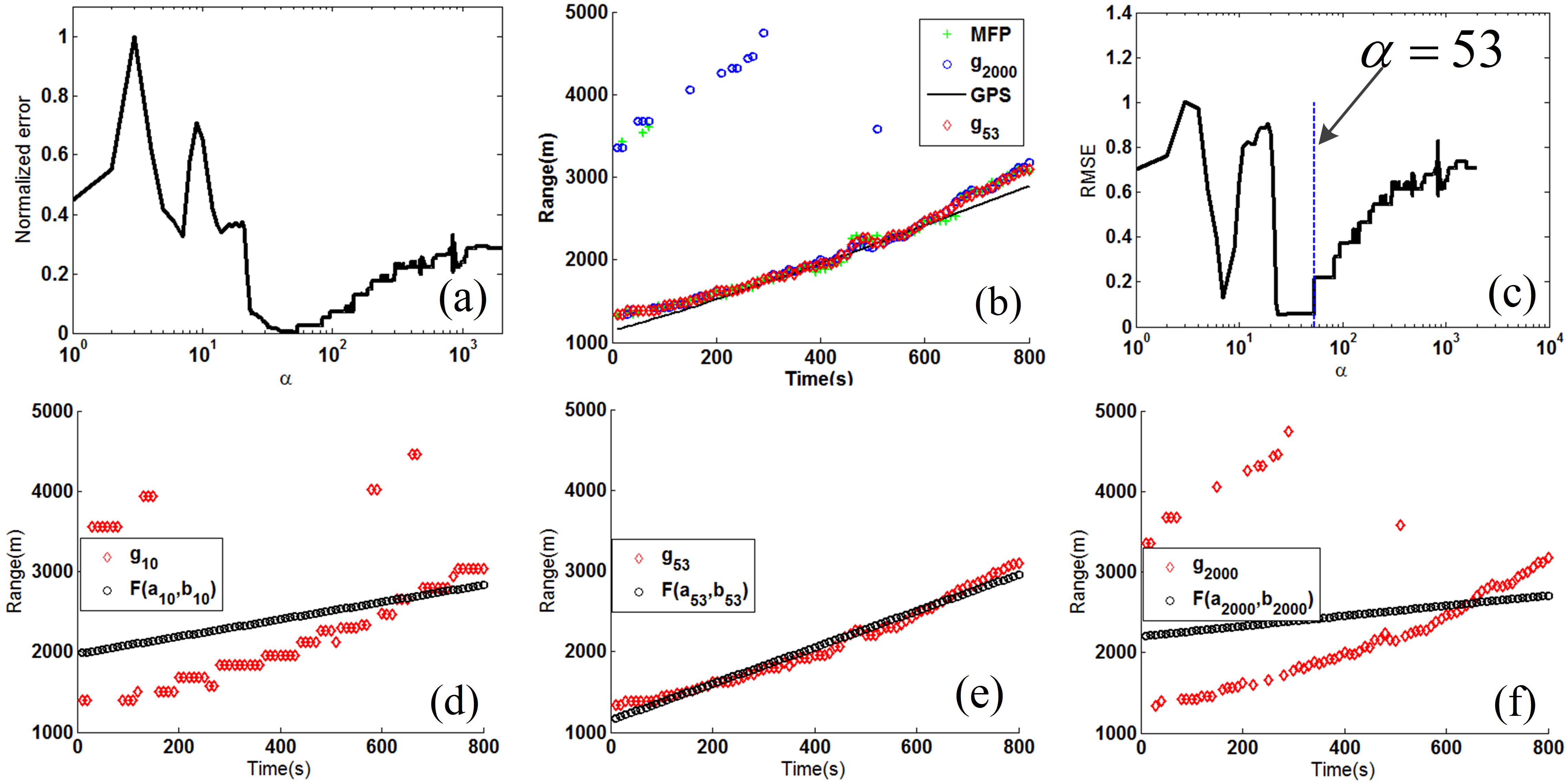}
\caption{(Color online)\label{fig3} Experimental data. {(a) Loss function $L_{\mathrm{FEAST}}(\alpha)$ is minimum when $\alpha=52$. (b) Range results $g_{53}(\hat{\mathbf{x}}_i)$ (diamonds), $g_{2000}(\hat{\mathbf{x}}_i)$ (circles) and MFP (crosses), and range from GPS (solid line). (c) RMSE for $g_\alpha(\hat{\mathbf{x}}_i)$.
$g_{\alpha}(\hat{\mathbf{x}}_i)$ (diamonds) and $F(\hat{\mathbf{x}}_i,a_\alpha,b_\alpha)$ (circles) for (d) $\alpha=10$, (e) $\alpha=52$ and (f) $\alpha=2000$.}}
\end{figure}

FEAST is demonstrated with the experimental data from the SWellEx-96 Event S5 \cite{website_12}. Only the 232 Hz shallow source that was towed at a depth of about 9 m is considered. The data recorded by VLA from $3700$ to $4500$ s are selected to prepare the experimental test data; every 10 s of data is used to construct a test data. The experimental test set contains 80 samples. Fig. \ref{fig3} (a) shows $L_{\mathrm{FEAST}}(\alpha)$ which is computed by the experimental test data and reaches the minimum at $\alpha=53$. Fig. \ref{fig3} (d)-(f) show
$g_\alpha(\hat{\mathbf{x}}_i)$ and $F(\hat{\mathbf{x}}_i,a_\alpha,b_\alpha)$ at different $\alpha$, and one again finds that $g_\alpha(\hat{\mathbf{x}}_i)$ and $F(\hat{\mathbf{x}}_i,a_\alpha,b_\alpha)$ are similar when $L_\mathrm{FEAST}(\alpha)$ reaches the minimum. Fig. \ref{fig3} (b) indicates that $g_{53}(\hat{\mathbf{x}}_i)$ is close to the GPS range. For comparison, Fig. \ref{fig3} (b) also shows the ranging results $g_{2000}(\hat{\mathbf{x}}_i)$ and the range from MFP where the ocean waveguide environment used in MFP is E1. It can be seen that, except for the points at 20, 60, 70 s, the range from MFP and $g_{53}(\hat{\mathbf{x}}_i)$ are almost the same and slightly deviate from the GPS range of the moving source, which is caused by the difference between E1 and the experimental environment. However, the ranging results $g_{2000}(\hat{\mathbf{x}}_i)$ have larger derivations from the GSP range. Fig. \ref{fig3} (c) gives the RMSE of $g_{\alpha}(\hat{\mathbf{x}}_i)$. One can find that when $\alpha=53$, the value of RMSE (0.0577) is closed to the minimum (0.0528), which verifies the FEAST again.

\section{\label{sec:5} Conclusion}

A method called FEAST is introduced to evaluate the ranging error of a FNN for source ranging on test data set. The FEAST is demonstrated by simulated and experimental data. FEAST, which requires that the trajectory of a moving sound source satisfies a known parameterized function, is used for data post-processing but not real-time processing.
The results indicates that FEAST improves the ranging accuracy of the FNN on test data. The FEAST is used for source ranging in this paper, but it can be used in other applications which has a known parameterized function.
\begin{acknowledgments}
This work is supported by the National Natural Science Foundation of China under Grant Nos. 11674294 and 11704359, the Fundamental Research Funds for the Central Universities under Grant No. 201861011 and Qingdao National Laboratory for Marine Science and Technology Foundation under Grant No. QNLM2016ORP0106. The authors also thank Ning Wang and Ruichun Tang for their useful suggestions for this paper.
\end{acknowledgments}


\newpage

\end{document}